\documentclass[journal]{IEEEtran}

\ifCLASSINFOpdf
\else
   \usepackage[dvips]{graphicx}
\fi
\usepackage{url}
\usepackage{amsmath}
\usepackage{amsfonts}
\usepackage{dsfont}
\usepackage{multirow}
\usepackage{subcaption}
\usepackage{booktabs}
\usepackage{makecell}
\usepackage{tabularx}
\usepackage{bbding}

\usepackage{framed}
\usepackage{latexsym}
\usepackage{color}
\usepackage{siunitx}
\usepackage{url}
\usepackage{xcolor}

\newlength\savewidth\newcommand\shline{\noalign{\global\savewidth\arrayrulewidth
  \global\arrayrulewidth 1pt}\hline\noalign{\global\arrayrulewidth\savewidth}}

\usepackage{hyperref}
\hypersetup{hypertex=true,
            colorlinks=true,
            linkcolor=red,
            anchorcolor=blue,
            citecolor=blue}

\usepackage{graphicx}

\begin{document}

\title{Sequential Visual and Semantic Consistency for Semi-supervised Text Recognition}

\author{Mingkun Yang, Biao Yang, Minghui Liao, Yingying Zhu*, and Xiang Bai, \IEEEmembership{Senior Member, IEEE}
\thanks{Mingkun Yang and Yingying Zhu are with the School of Electronic Information and Communications, Huazhong University of Science and Technology, Wuhan 430074, China. Yingying Zhu is also with the Hubei Key Laboratory of Smart Internet Technology, Huazhong University of Science and Technology, Wuhan 430074, China. (e-mail: yangmingkun@hust.edu.cn, yyzhu@hust.edu.cn). \textit{(Corresponding author: Yingying Zhu.)}}
\thanks{Biao Yang is with the School of Artificial Intelligence and Automation, Huazhong University of Science and Technology, Wuhan 430074, China (e-mail: hust\_byang@hust.edu.cn, xbai@hust.edu.cn).}
\thanks{Xiang Bai is with the School of Software Engineering,  Huazhong University of Science and Technology, Wuhan 430074, China (e-mail: xbai@hust.edu.cn).}
\thanks{Minghui Liao is with the Huawei Cloud, Shenzhen 518129, China (e-mail: mhliao@foxmail.com).}
}

\maketitle

\begin{abstract}
    Scene text recognition (STR) is a challenging task that requires large-scale annotated data for training. However, collecting and labeling real text images is expensive and time-consuming, which limits the availability of real data. Therefore, most existing STR methods resort to synthetic data, which may introduce domain discrepancy and degrade the performance of STR models. To alleviate this problem, recent semi-supervised STR methods exploit unlabeled real data by enforcing character-level consistency regularization between weakly and strongly augmented views of the same image. However, these methods neglect word-level consistency, which is crucial for sequence recognition tasks. This paper proposes a novel semi-supervised learning method for STR that incorporates word-level consistency regularization from both visual and semantic aspects. Specifically, we devise a shortest path alignment module to align the sequential visual features of different views and minimize their distance. Moreover, we adopt a reinforcement learning framework to optimize the semantic similarity of the predicted strings in the embedding space. We conduct extensive experiments on several standard and challenging STR benchmarks and demonstrate the superiority of our proposed method over existing semi-supervised STR methods.
\end{abstract}

\begin{IEEEkeywords}
Semi-supervised Learning, Scene Text Recognition, Dynamic Programming, Reinforcement Learning
\end{IEEEkeywords}

\IEEEpeerreviewmaketitle

\section{Introduction}
\label{sec:introduction}

\IEEEPARstart{T}{ext} understanding from natural images is a fundamental and active research topic in both vision and language fields, as it has numerous potential applications in various domains, such as autonomous driving~\cite{zhu2017cascaded}, digital financial system, and product search~\cite{DBLP:journals/access/BaiYLXL18}. Scene text recognition (STR), which aims to transcribe text content from images, is an essential component of the text understanding system. However, STR is still challenging due to the complexity and diversity of natural scenes, such as cluttered backgrounds, arbitrary font styles, distorted shapes, and occlusions.

Most existing STR methods require substantial quantities of training data. However, acquiring and labeling real images can be both costly and time-intensive. Consequently, these methods often rely on large-scale synthetic training data, such as Synth90k~\cite{synth90k} and SynthText~\cite{synthtext}, which can yield satisfactory results on standard benchmarks. Nevertheless, synthetic data cannot encompass all situations where text appears, and this domain gap between synthetic and real data can restrict the performance of STR models in realistic settings. One promising solution is to incorporate unlabeled real images with labeled synthetic images, which could improve the performance of STR models and facilitate their application in practical scenarios.

Semi-supervised learning (SSL) is a natural way to exploit both unlabeled real data and labeled synthetic data for STR. Baek~\textit{et al.}~\cite{baek2021if} and Fang~\textit{et al.}~\cite{abinet} adopted the self-training framework for STR. In this framework, the STR model was first trained in a supervised way. Then, the model was used to generate pseudo labels for unlabeled data, which were subsequently applied to retrain the model. The separate steps involved in this framework can affect the training efficiency. To overcome this limitation, Zheng~\textit{et al.}~\cite{zheng2022pushing} employed consistency regularization (CR), another popular SSL technique, to STR. The CR-based SSL methods enforced consistency between the predictions of two views of the same unlabeled image. Due to the online training process, CR-based methods usually performed better than self-training on several SSL benchmarks~\cite{abuduweili2021adaptive,RizveDRS21semi} and were more efficient. However, Zheng~\textit{et al.} observed that directly applying the CR-based methods, such as UDA~\cite{conf/nips/XieDHL020} and FixMatch~\cite{sohn2020fixmatch}, was not suitable for semi-supervised STR. They attributed this to the misalignment between character sequences, which led to an unstable training process. Therefore, they proposed character-level consistency regularization (CCR) to ensure better sequence alignment between predictions from different views. This was achieved by using the teacher forcing strategy, in which the predicted character from a previous time step of the teacher model was used to train the student model, rather than relying on the student model’s prediction.

Text is sequential data composed of characters, which requires special consideration in SSL. We propose word-level consistency regularization from both visual and semantic aspects to account for the sequential nature of text. CCR can enforce character-level consistency constraints by using teacher forcing, but this may not be reliable. If the teacher model’s output is erroneous, one-to-one character matching may result in misalignment between the teacher and student model’s output sequences. To address this problem, we devise the shortest path loss between two sequential glimpse vectors via dynamic programming to enforce word-level visual consistency regularization. Moreover, some symbols are difficult to distinguish visually, which leads to small differences between their visual features and no significant penalty in the semi-supervised optimization. However, when considering the semantics of a whole word, two words with similar characters may have very different meanings. Therefore, we introduce word-level semantic consistency by aligning the embeddings of predicted words in a semantic space. Since the word embedding is derived from predicted words, it is non-differentiable to optimize the semantic similarity directly. To overcome this challenge, we employ reinforcement learning, inspired by self-critical image captioning~\cite{conf/cvpr/RennieMMRG17}, to measure semantic consistency and make the whole pipeline differentiable and trainable.

The main contributions of this paper can be summarized as follows,
\begin{enumerate}
  \item We propose the shortest path loss via dynamic programming to enforce word-level visual consistency, which can effectively mitigate the sequence unalignment issue caused by incorrect prediction of the teacher model.
  \item We introduce word-level semantic consistency regularization and employ reinforcement learning for differentiable training, which can distinguish similar characters from the semantic aspect.
  \item We combine word-level and character-level CR and achieve state-of-the-art performance on standard benchmarks and more challenging datasets.
\end{enumerate}

\section{Related Works}
\label{sec:related_work}

\subsection{Scene Text Recognition}
Drawing inspiration from works on automatic speech recognition and machine translation, STR can be viewed as a sequence recognition task, where the text image is initially encoded into a feature sequence and subsequently decoded as a character sequence. In \cite{baek2019wrong}, Beak~\textit{et al.}~modularized the STR framework into four components, i.e., \textit{Transformation}, \textit{Feature Extraction}, \textit{Sequence Modeling}, and \textit{Prediction}. 
While several works~\cite{aster,cluo2019moran,YangGLHBBYB19} primarily focus on the \textit{Transformation} component for irregular text recognition, the \textit{Prediction} component is the most explored part in the STR field~\cite{crnn,zhang2020sahan,zhang2021pmmn}.
Recently, motivated by the Levenshtein Transformer decoder, Da~\textit{et al.}~\cite{DBLP:conf/eccv/DaWY22levocr} cast STR as an iterative sequence refinement procedure, and devised to delete or insert characters to refine the preceding predictions dynamically. To integrate linguistic knowledge into STR, Wang~\textit{et al.}~\cite{DBLP:conf/eccv/WangDY22multigra} introduced a Multi-Granularity Prediction strategy to inject linguistic information through additionally predicting BPE and WordPiece. ABINet~\cite{abinet} employed a standalone Language Model for prediction refinement, while PARSeq~\cite{DBLP:conf/eccv/BautistaA22parseq} presented a more efficient approach using permuted auto-regressive sequence decoder. Owing to its exceptional performance and flexibility, attention-based predictor (LSTM or Transformer) has been dominant in the STR field.

\begin{figure*}[t]
\centering
\includegraphics[width=0.98\linewidth]{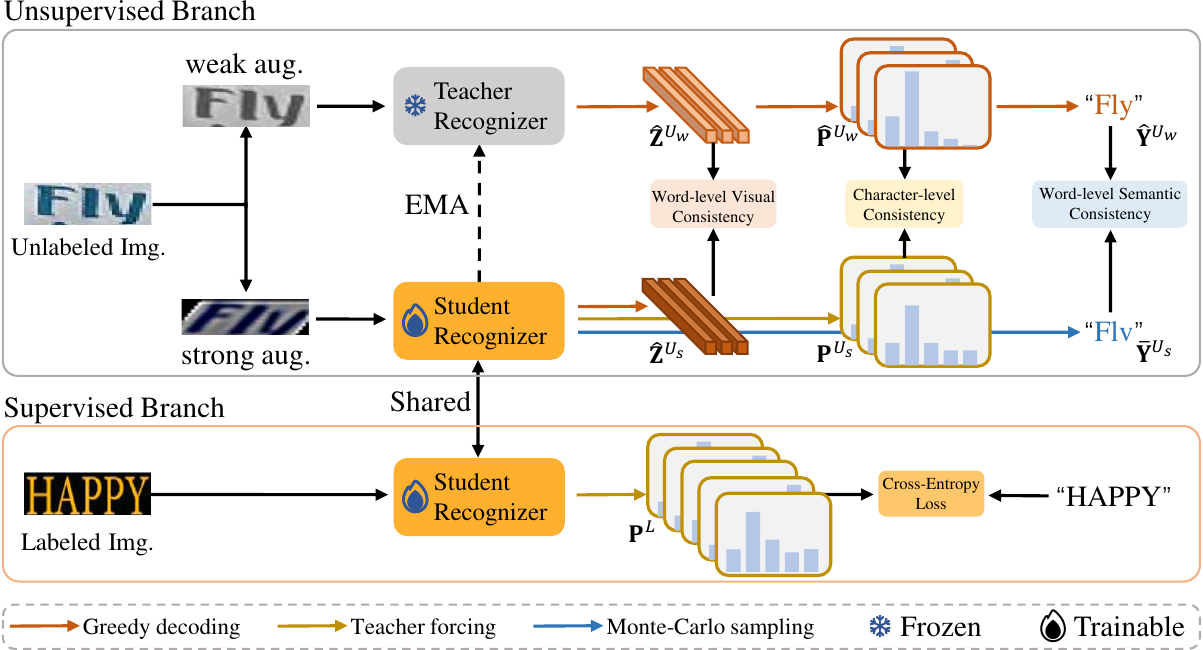}
\caption{Overall pipeline of our proposed method. A regular recognizer is trained in a supervised manner with labeled synthetic images. A Mean-Teacher architecture is employed for unlabeled real data to conduct unsupervised learning via consistency regularization (CR). Besides the CCR on the character-level probability distribution, word-level visual CR on the glimpse vectors and semantic CR on the predicted strings are introduced. Best viewed in color.}
\label{fig:methodology}
\end{figure*}

\subsection{Semi-supervised Learning}
Semi-supervised learning (SSL) is a machine learning approach that addresses scenarios where obtaining labeled data is expensive, while unlabeled data is abundant. By utilizing a small amount of labeled data in conjunction with unlabeled data, SSL can significantly improve learning accuracy. Two prominent families of SSL are self-training and consistency regularization-based methods.
Self-training involves training a model with labeled data and using it to produce pseudo labels for unlabeled data, which are then utilized to retrain the model. However, the separate steps involved in this pipeline may affect training efficiency. CR-based methods share a similar viewpoint with discriminative self-supervised methods and aim to ensure that the representations of different augmentation views from the same images are consistent.
Mean Teacher~\cite{DBLP:conf/nips/TarvainenV17meanteacher} was a CR-based method that enforces consistency between the probability distributions of the student model and teacher model. To prevent learning collapse, the teacher model was updated through exponential moving averaging of the student model. Benefiting from both CR and pseudo-labeling, FixMatch~\cite{sohn2020fixmatch} produced better performance. By introducing perturbation into consistency training, UDA~\cite{conf/nips/XieDHL020} performed much better on various tasks.

\subsection{Semi-supervised Text Recognition}

While numerous SSL methods have been proposed for image classification~\cite{boschini2022continual}, semantic segmentation~\cite{feng2023weakly}, and object detection~\cite{an2023prouda}, there is a scarcity of works focusing on SSL for STR. Baek~\textit{et al.}~\cite{baek2021if} attempted to enhance STR performance using Pseudo Label or Mean Teacher with labeled real data. Fang~\textit{et al.}~\cite{abinet} proposed an ensemble self-training strategy to further increase the performance of their proposed ABINet. Gaurav~\textit{et al.}~\cite{patel2023seq} utilized Beam-Search inference for pseudo-label assignment, and a character and sequence aware uncertainty estimate for sample selection. However, these self-training methods~\cite{baek2021if, abinet, patel2023seq} require iterative model training with labeled data and the generated pseudo labels for unlabeled data, resulting in a separate pipeline. This not only impacts training efficiency but also leads to an unstable training process, particularly when there is a significant domain gap between the labeled and unlabeled data~\cite{zheng2022pushing}. To overcome this limitation, Zheng~\textit{et al.}~\cite{zheng2022pushing} proposed character-level consistency regularization to mitigate the training instability. However, their approach does not explore the misalignment caused by incorrect predictions or leverage semantic information of the predicted text. This oversight may restrict the performance of SSL for STR.

\section{Methodology}
\label{sec:methodology}

\subsection{Preliminaries}
\label{subsec:preliminaries}

Based on the state-of-the-art semi-supervised STR method~\cite{zheng2022pushing}, we adopt a Mean-Teacher architecture~\cite{DBLP:conf/nips/TarvainenV17meanteacher} with CCR as our baseline.
To ensure efficiency and performance, the STR model comprises ConvNextv2-based~\cite{woo2023convnext} encoder and transformer-based decoder~\cite{DBLP:conf/nips/VaswaniSPUJGKP17attn}. The whole pipeline is illustrated in Fig.~\ref{fig:methodology}. For clarity, we briefly introduce two strategies for sequence decoding: teacher forcing, where the ground-truth character is fed into the decoder for predicting the next character, and greedy decoding, where the decoder uses its own prediction from the previous step.

Given a labeled synthetic image $\mathbf{X}^L$ and its ground-truth character sequence $\mathbf{Y}^L=\{y_1^L,y_2^L,...,y_T^L\}$, the student STR model produces a sequence of probability distribution $\mathbf{P}^L=\{p_1^L,p_2^L,...,p_T^L\}$ via \textit{Softmax} and a sequence of glimpse vectors $\mathbf{Z}^L=\{z_1^L,z_2^L,...,z_T^L\}$. The student model is trained by minimizing the cross-entropy loss $\mathcal{L}_{ce}$ on $\mathbf{P}^L$.

An unlabeled image $\mathbf{X}^U$ first goes through a weak and strong augmentation, resulting in two augmented views, $\mathbf{X}^{U_w}$ and $\mathbf{X}^{U_s}$. The teacher model takes $\mathbf{X}^{U_w}$ as input to generates pseudo labels $\hat{\mathbf{Y}}^{U_w}$ and probability distributions $\hat{\mathbf{P}}^{U_w}$ with length $T_w$ by greedy decoding. In CCR, $\hat{\mathbf{Y}}^{U_w}$ and $\mathbf{X}^{U_s}$ are then used to yield $\mathbf{P}^{U_s}$ with teacher forcing. To filter out noise unlabeled samples, a threshold $\tau$ on confidence score $\hat{\mathbf{S}}^{U_w}$ is applied for student training. The confidence score is the cumulative product of the maximum output probability of each decoding step $\hat{p}^{U_w}_i$. The character-level consistency regularization in the unsupervised branch is defined as,

\begin{equation}
    \mathcal{L}_{ccr} = \mathds{1}(\hat{\mathbf{S}}^{U_w} > \tau) \frac{1}{T_w}\sum_i^{T_w}KLDiv(\hat{p}_i^{U_w}, p_i^{U_s})
\label{eqn:loss_ccr}
\end{equation}

\noindent where $\mathds{1}(\cdot)$ is an indicator function. $KLDiv(\cdot)$ is KL-divergence which is used to align two probability distributions.

Finally, we maintain a teacher STR model $f_t(\cdot;\Theta_t)$ and a student model $f_s(\cdot;\Theta_s)$ that minimize the loss

\begin{equation}
    \mathcal{L} = \mathcal{L}_{ce} + \lambda_1\mathcal{L}_{ccr},
\end{equation}

\noindent where $\lambda_1$ is a weighting parameter. The teacher model is an exponential moving average (EMA) of the student STR model, which is updated as $\Theta_t \leftarrow (1-\gamma) \Theta_t + \gamma \Theta_s$.

\begin{figure}[t]
\centering
\includegraphics[width=0.99\linewidth]{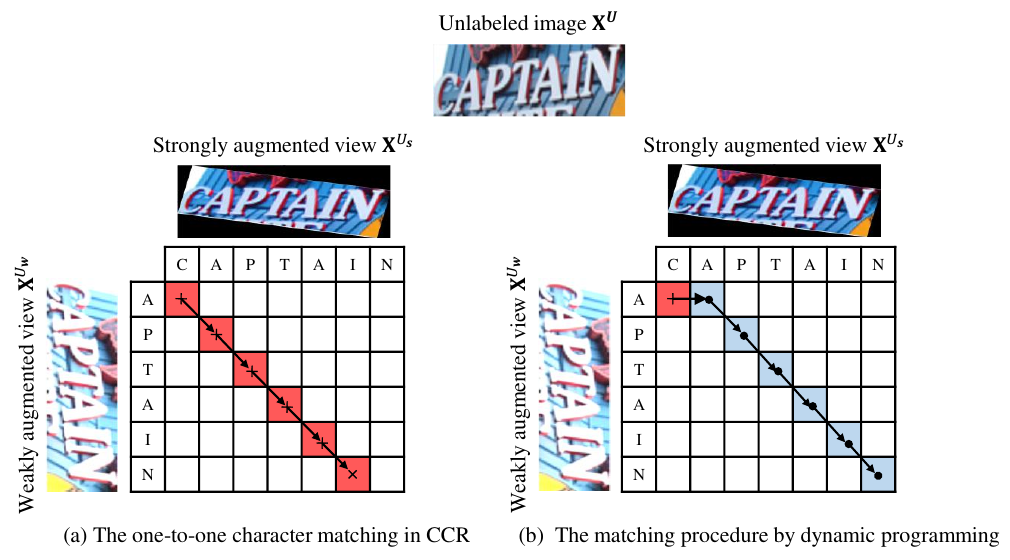}
\caption{The comparison between the word-level consistencies calculated via one-to-one character matching and dynamic programming. The characters displayed at the left and top of the grid represent the predicted strings from \textit{Teacher Recognizer} and \textit{Student Recognizer}, respectively. In the grid, the red block indicates an incorrect matching while an azure block represents a correct matching.}
\label{fig:matching_comparison}
\end{figure}

\subsection{Word-level Visual Consistency}
\label{subsec:seq_vis_cons}
CCR enforces character-level consistency regularization by treating the pseudo labels $\hat{\mathbf{Y}}^{U_w}$ as the ground-truth of $\mathbf{X}^{U_s}$ and using teacher forcing strategy to ensure alignment between the sequences $\hat{\mathbf{P}}^{U_w}$ and $\mathbf{P}^{U_s}$. However, some noise samples may still exist after applying a threshold, due to the overconfidence issue~\cite{DBLP:conf/icml/GuoPSW17calib}. 
If the teacher model’s prediction is incorrect, the one-to-one character matching may cause misalignment and thus hinder the training process. Therefore, we introduce a sequential consistency regularization at the word level. Specifically, we first obtain $\hat{\mathbf{Z}}^{U_w}$ from Sec.~\ref{subsec:preliminaries}. Then, instead of using teacher forcing to forward $\mathbf{X}^{U_s}$, we also employ greedy decoding to obtain glimpse vectors $\hat{\mathbf{Z}}^{U_s}$ with length $T_s$, avoiding incorrect prediction from the teacher model. As shown in Fig.~\ref{fig:matching_comparison}, without teacher forcing, $T_w$ and $T_s$ may be different. Thus, we need to measure the consistency between two sequences with different lengths. We do this by dynamically matching the two sequential vectors to find the alignment of glimpse vectors with the minimum total distance. 

During greedy decoding, obtaining predictions from previous time steps requires a non-differentiable \textit{argmax} operation since characters are discrete tokens. This results in inefficiencies when modeling the influence of prior time steps on subsequent predictions. To address this issue, we take inspiration from~\cite{gu2018neural}, where the straight-through version of the Gumbel-Softmax estimator, namely \textit{ST-Gumbel}~\cite{DBLP:conf/iclr/JangGP17} was utilized to generate differentiable sequences of discrete words for neural machine translation. In our approach, we use the \textit{ST-Gumbel} operation instead of \textit{Softmax+argmax} at each time step for character prediction, allowing for efficient sequence modeling during greedy decoding while bypassing non-differentiability.

With $\hat{\mathbf{Z}}^{U_w}=\{\hat{z}_1^{U_w},\hat{z}_2^{U_w},...,\hat{z}_{T_w}^{U_w}\}$ and $\hat{\mathbf{Z}}^{U_s}=\{\hat{z}_1^{U_s},\hat{z}_2^{U_s},...,\hat{z}_{T_s}^{U_s}\}$, we first calculate the \textit{cosine} distance $d_{ij}$ between $\hat{z}_i^{U_w}$ and $\hat{z}_j^{U_s}$, resulting in a distance matrix $D\in\mathbb{R}^{T_w \times T_s}$. We define the word-level visual consistency regularization as minimizing the total distance of the shortest path from $(1,1)$ to $(T_w, T_s)$ in the matrix $D$ and employ dynamic programming to calculate the distance as follows,

\small
\begin{equation}\label{ShortestPath}
S_{i,j}=\begin{cases}
d_{ij} & i=1,  j=1 \\
S_{i-1,j}+d_{ij} & i \ne 1,j=1 \\
S_{i,j-1}+d_{ij} & i=1,j \ne 1 \\
\text{min}(S_{i-1,j},S_{i,j-1},S_{i-1,j-1})+d_{ij} & i \ne 1, j \ne 1,
\end{cases}
\end{equation}

\noindent where $S_{i,j}$ is the shortest path distance from $(1,1)$ to $(i,j)$ and $S_{T_w,T_s}$ is final shortest distance between $\hat{\mathbf{Z}}^{U_w}$ and $\hat{\mathbf{Z}}^{U_s}$. By additionally minimizing $S_{T_w,T_s}$, the sequential CR can alleviate the misalignment caused by the false positives in CCR.

\subsection{Word-level Semantic Consistency}
\label{subsec:seq_sem_cons}
For text recognition, some characters may be visually similar in complex scenarios, such as \{`o', `0'\}, \{`1', `l', `i'\}, \{`P', `R', `k', `B'\}, and so on. The similarity between these characters can result in minor differences between their visual features, and thus no significant penalty during semi-supervised learning. However, these visually similar characters can be easily distinguished from the word-level semantics. Therefore, we introduce word-level semantic consistency regularization to help the student model learn more discriminative representations.

The word-level semantics are represented based on the predicted words, making it non-differentiable to optimize the semantic similarity directly.
Inspired by~\cite{conf/cvpr/RennieMMRG17}, which optimizes the non-differentiable NLP metrics via Self-critical Sequence Training (SCST), a variant of the popular REINFORCE algorithm, we also use SCST to constraint the semantic consistency between the teacher and student models. More details about the derivation process of SCST can be found in \cite{conf/cvpr/RennieMMRG17}. In summary, the goal of SCST is to minimize the negative expected reward, which is computed through the following objective,

\begin{equation}
    \mathcal{L}_{scst} = - (r(\bar{\mathbf{Y}}^{U_s}, \hat{\mathbf{Y}}^{U_w}) - r(\hat{\mathbf{Y}}^{U_s}, \hat{\mathbf{Y}}^{U_w}))\frac{1}{\bar{T}_s}\sum_t^{\bar{T}_s}\log \bar{p}_t^{U_s}
\end{equation}

\noindent where $r(\cdot, \cdot)$ denotes the reward and is computed by \textit{cosine} similarity between the word embeddings of the generated string from the student model and the corresponding pseudo label, $\hat{\mathbf{Y}}^{U_w}$ from the teacher model. Specifically, we employ fastText~\cite{bojanowski2017enriching} to embed the predicted words into a continuous semantic space. In fastText, each character n-gram has a vector representation, and the words are represented as the sum of these representations. Therefore, fastText can handle the out-of-vocabulary (OOV) issue, which is common in practical STR applications. $\bar{\mathbf{Y}}^{U_s}=\{\bar{y}_1^{U_s},\bar{y}_2^{U_s},...,\bar{y}_{\bar{T}_s}^{U_s}\}$ and $\bar{y}_t^{U_s}$ is the character obtained by a single \textit{Monte-Carlo} sampling from the probability distribution $\bar{p}_t^{U_s}$ at time step $t$. 
$\bar{p}_t^{U_s}$ is produced via greedy decoding where the previous character is $\bar{y}_{t-1}^{U_s}$.
$\hat{\mathbf{Y}}^{U_s}$ represents the predicted character sequence obtained via greedy decoding with \textit{Softmax} and used as the ``baseline''. It is the same as the pseudo labels $\hat{\mathbf{Y}}^{U_w}$ in Sec.~\ref{subsec:seq_vis_cons} but with different input images, namely $U_s$ and $U_w$.

In SCST, the ``baseline'' is obtained via greedy decoding, which is consistent with the test-time inference procedure. Compared with other REINFORCE algorithms, SCST can reduce variance for more effective training and force improve the model's performance under the inference algorithm used at test time.

Through SCST, sampled words from the student model that return higher rewards than $\hat{\mathbf{Y}}^{U_s}$ are ``pushed up" by increasing their probability, while samples that result in lower reward are suppressed. This process encourages the student model to generate words that are more semantically consistent with the teacher model.

\begin{table}[b]
\centering
\caption{Details of Scene Text Recognition Benchmarks.}
\label{tab:benchmarks}
\resizebox{0.45\textwidth}{!}{
\begin{tabular}{lcc}
\toprule
Benchmarks                                 & \# Test images    & Description \\ \midrule
IIIT~\cite{DBLP:conf/cvpr/MishraAJ12}      &   3000            &  Regular Scene Text           \\ 
SVT~\cite{wang2011end}                     &   647             &  Regular Scene Text          \\ 
IC13~\cite{karatzas2013icdar}              &   1015            &  Regular Scene Text          \\ \midrule
IC15~\cite{karatzas2015icdar}              &   1811            &  Incidental Scene Text          \\
COCO~\cite{cocotext}                       &   9896            &  Incidental Scene Text          \\ \midrule
SVTP~\cite{quy2013recognizing}             &   645             &  Perspective Scene Text         \\ \midrule
CUTE~\cite{risnumawan2014robust}           &   288             &  Curved Scene Text         \\ 
CTW~\cite{ctw1500}                         &   1572            &  Curved Scene Text         \\ 
TT~\cite{CK2017}                           &   2201            &  Curved Scene Text        \\ \midrule
WOST~\cite{ost}                            &   2416            &  Weakly Occluded Scene Text         \\ 
HOST~\cite{ost}                            &   2416            &  Heavily Occluded Scene Text         \\ \bottomrule
\end{tabular}}
\end{table}

\begin{table}[]
\centering
\caption{Detailed training settings.}
\label{tab:optim_hyperparams}
\begin{tabular}{l|l}
Config                       & Value               \\ \shline
optimizer                    & AdamW               \\
base learning rate           & 8e-4                \\
weight decay                 & 0.05                \\
optimizer momentum           & $\beta_1$, $\beta_2$ = 0.9, 0.999 \\
batch size                   & 1,024               \\
learning rate schedule       & \textit{cosine} decay        \\
warmup epochs                & 1                   \\
training epochs              & 10                  \\
unlabeled/labeled data ratio & 1:1                 \\
teacher decay rate $\gamma$         & 0.9999              \\
confidence threshold  $\tau$        & 0.5                 \\
loss weights $\lambda_1$, $\lambda_2$, and $\lambda_3$      & 1, 0.1, 0.1        
\end{tabular}
\end{table}

\subsection{Overall objective}
\label{subsec:overall_obj}
Based on the CCR, we additionally introduce two word-level CRs from visual and semantic perspectives. Therefore, the hierarchical loss function benefits from the complementary of both character-level and word-level, as well as visual and semantics. Both labeled data and unlabeled data are trained together to optimize the overall objective function,

\begin{equation}
    \mathcal{L} = \mathcal{L}_{ce} + \lambda_1\mathcal{L}_{ccr} + \lambda_2S_{T_w,T_s} + \lambda_3\mathcal{L}_{scst},
\label{eqn:overall_obj}
\end{equation}

\noindent where $\lambda_1$, $\lambda_2$, $\lambda_3$ are hyper-parameters to balance different losses.

\section{Experiments}
\label{sec:experiments}

\begin{table*}[!ht]
\caption{Scene text recognition results on standard benchmarks. ``*" indicates the results are tested with the officially released codes and models. \textbf{Bold} represents the best performance. \underline{Underline} denotes the second best result. ``WiKi" indicates using a language model trained with data from WiKiText-103~\cite{wiki}.}
\label{tab:sota-tr}
\resizebox{1.\textwidth}{!}{
\begin{tabular}{ccccccccccccccccc}
\toprule
\multicolumn{1}{c}{\multirow{2}{*}{Method}} & \multicolumn{1}{c}{\multirow{2}{*}{Venue}} & \multicolumn{1}{c}{\multirow{2}{*}{\begin{tabular}[c]{@{}c@{}}Labeled\\ Data\end{tabular}}} & \multicolumn{1}{c}{\multirow{2}{*}{\begin{tabular}[c]{@{}c@{}}Unlabeled\\ Data\end{tabular}}} & \multicolumn{3}{c}{Regular} & \multicolumn{6}{c}{Irregular}           & \multicolumn{2}{c}{Occluded}     & \multirow{2}{*}{Avg.}     & \multirow{2}{*}{Params.} \\
\cmidrule(lr){5-7} \cmidrule(lr){8-13} \cmidrule(lr){14-15}
\multicolumn{1}{c}{}     & \multicolumn{1}{c}{}    & \multicolumn{1}{c}{}  & \multicolumn{1}{c}{}                                     & IIIT    & SVT     & IC13    & IC15 & SVTP & CUTE & COCO & CTW  & TT   & HOST & WOST                      &                           &                          \\ \midrule
\multicolumn{1}{l|}{Mask TextSpotter~\cite{LiaoLHYWB21}} & \multicolumn{1}{c|}{TPAMI2021} & \multicolumn{1}{c|}{STD}    & \multicolumn{1}{c|}{-}                       & 95.3    & 91.8    & 95.3    & 78.2 & 83.6 & 88.5 & -    & -    & -    & -    & \multicolumn{1}{c|}{-}    & \multicolumn{1}{c|}{-}    & -                     \\
\multicolumn{1}{l|}{PREN2D~\cite{Pren2D}}   & \multicolumn{1}{c|}{CVPR2021}              & \multicolumn{1}{c|}{STD}   & \multicolumn{1}{c|}{-}                       & 95.6    & 94      & 96.4    & 83   & 87.6 & 91.7 & -    & -    & -    & -    & \multicolumn{1}{c|}{-}    & \multicolumn{1}{c|}{-}    & -                        \\
\multicolumn{1}{l|}{ABINet~\cite{abinet}}   & \multicolumn{1}{c|}{CVPR2021}          & \multicolumn{1}{c|}{STD+WiKi}  & \multicolumn{1}{c|}{-}                       & 96.2    & 93.5    & 97.4    & 86.0 & 89.3 & 89.2 & -    & -    & -    & -    & \multicolumn{1}{c|}{-}    & \multicolumn{1}{c|}{-}    & 37M                      \\
\multicolumn{1}{l|}{ABINet*}                & \multicolumn{1}{c|}{CVPR2021} & \multicolumn{1}{c|}{STD+WiKi}            & \multicolumn{1}{c|}{-}                       & 96.4    & 94.4    & 97.0    & 85.9 & 89.6 & 88.5 & 63.0 & 76.8 & 80.7 & 57.9 & \multicolumn{1}{c|}{75.3} & \multicolumn{1}{c|}{74.4} & 37M                      \\
\multicolumn{1}{l|}{SATRN~\cite{satrn}}   & \multicolumn{1}{c|}{CVPRW2021}              & \multicolumn{1}{c|}{STD}    & \multicolumn{1}{c|}{-}                       & 92.8    & 91.3    & -       & -    & 86.5 & 87.8 & -    & -    & -    & -    & \multicolumn{1}{c|}{-}    & \multicolumn{1}{c|}{-}    & 55M                      \\
\multicolumn{1}{l|}{SATRN*}               & \multicolumn{1}{c|}{CVPRW2021}  & \multicolumn{1}{c|}{STD}                 & \multicolumn{1}{c|}{-}                       & 92.5    & 93.9    & 96.3    & 81.3 & 86.5 & 85.8 & 50.4 & 70.8 & 74   & 63.4 & \multicolumn{1}{c|}{76.3} & \multicolumn{1}{c|}{68.2} & 55M                      \\
\multicolumn{1}{l|}{VisionLAN~\cite{ost}}   & \multicolumn{1}{c|}{ICCV2021}           & \multicolumn{1}{c|}{STD}      & \multicolumn{1}{c|}{-}                       & 95.8    & 91.7    & 95.7    & 83.7 & 86.0 & 88.5 & -    & -    & -    & -    & \multicolumn{1}{c|}{-}    & \multicolumn{1}{c|}{-}    & 33M                      \\
\multicolumn{1}{l|}{VisionLAN*}             & \multicolumn{1}{c|}{ICCV2021} & \multicolumn{1}{c|}{STD}                 & \multicolumn{1}{c|}{-}                       & 95.9    & 92.0    & 96.3    & 84.1 & 85.9 & 88.9 & 59.2 & 75.1 & 78.7 & 49.8 & \multicolumn{1}{c|}{70.8} & \multicolumn{1}{c|}{71.1} & 33M                      \\
\multicolumn{1}{l|}{JVSR~\cite{JVSR}}    & \multicolumn{1}{c|}{ICCV2021}               & \multicolumn{1}{c|}{STD}     & \multicolumn{1}{c|}{-}                       & 95.2    & 92.2    & -       & -    & 85.7 & 89.7 & -    & -    & -    & -    & \multicolumn{1}{c|}{-}    & \multicolumn{1}{c|}{-}    & -                        \\
\multicolumn{1}{l|}{Text is Text~\cite{textistext}}    & \multicolumn{1}{c|}{ICCV2021}               & \multicolumn{1}{c|}{STD}     & \multicolumn{1}{c|}{-}                       & 92.3    & 89.9    & 93.3       & 76.9    & 84.4 & 86.3 & -    & -    & -    & -    & \multicolumn{1}{c|}{-}    & \multicolumn{1}{c|}{-}    & -                        \\
\multicolumn{1}{l|}{SVTR~\cite{DBLP:conf/ijcai/DuCJYZLDJ22svtr}}    & \multicolumn{1}{c|}{IJCAI2022}                      & \multicolumn{1}{c|}{STD}       & \multicolumn{1}{c|}{-}                       & 96.3    & 91.7    & 97.2    & 86.6 & 88.4 & 95.1 & -    & -    & -    & -    & \multicolumn{1}{c|}{-}    & \multicolumn{1}{c|}{-} & 41M                      \\ 
\multicolumn{1}{l|}{ConCLR~\cite{DBLP:conf/aaai/ZhangZYSL022conclr}}    & \multicolumn{1}{c|}{AAAI2022}                      & \multicolumn{1}{c|}{STD}       & \multicolumn{1}{c|}{-}                       & 95.7    & 92.1    & 95.9    & 84.4 & 85.7 & 89.2 & -    & -    & -    & -    & \multicolumn{1}{c|}{-}    & \multicolumn{1}{c|}{-} & -                      \\ 
\multicolumn{1}{l|}{S-GTR~\cite{DBLP:conf/aaai/HeC0LHWD22sgtr}}    & \multicolumn{1}{c|}{AAAI2022}                      & \multicolumn{1}{c|}{STD}       & \multicolumn{1}{c|}{-}                       & 96.8    & 94.8    & 97.7   & 86.9 & 89.6 & 93.1 & -    & -    & -    & -    & \multicolumn{1}{c|}{-}    & \multicolumn{1}{c|}{-} & 42M                      \\ 
\multicolumn{1}{l|}{CornerTransformer~\cite{DBLP:conf/eccv/XieFZWB22wordart}}  & \multicolumn{1}{c|}{ECCV2022}      & \multicolumn{1}{c|}{STD}       & \multicolumn{1}{c|}{-}                       & 95.9    & 94.6    & 96.4    & 86.3 & \underline{91.5} & 92.0  & -    & -    & -    & -    & \multicolumn{1}{c|}{-}   & \multicolumn{1}{c|}{-} & 86M                      \\
\multicolumn{1}{l|}{LevOCR~\cite{DBLP:conf/eccv/DaWY22levocr}}     & \multicolumn{1}{c|}{ECCV2022}              & \multicolumn{1}{c|}{STD}       & \multicolumn{1}{c|}{-}                       & 96.6    & 92.9    & 96.7    & 86.4 & 88.1 & 91.7 & -    & -    & -    & -    & \multicolumn{1}{c|}{-}    & \multicolumn{1}{c|}{-}    & 93M                     \\
\multicolumn{1}{l|}{$\text{MGP-STR}_{Fuse}$~\cite{DBLP:conf/eccv/WangDY22multigra}}    & \multicolumn{1}{c|}{ECCV2022}       & \multicolumn{1}{c|}{STD}       & \multicolumn{1}{c|}{-}                       & 96.4    & 94.7    & 97.3    & 87.2 & 91.0 & 90.3 & -    & -    & -    & -    & \multicolumn{1}{c|}{-}    & \multicolumn{1}{c|}{-} & 148M                      \\
\multicolumn{1}{l|}{PARSeq~\cite{DBLP:conf/eccv/BautistaA22parseq}}    & \multicolumn{1}{c|}{ECCV2022}               & \multicolumn{1}{c|}{STD}       & \multicolumn{1}{c|}{-}                       & 97.0    & 93.6    & 96.2    & 82.9 & 88.9 & 92.2 & -    & -    & -    & -    & \multicolumn{1}{c|}{-}    & \multicolumn{1}{c|}{-} & 24M                      \\ 
\multicolumn{1}{l|}{$\text{SIGA}_T$~\cite{guan2023self}}            & \multicolumn{1}{c|}{CVPR2023}    & \multicolumn{1}{c|}{STD}       & \multicolumn{1}{c|}{-}                       & 96.6    & 95.1    & 97.8    & 86.6 & 90.5 & 93.1 & - & - & - & - & \multicolumn{1}{c|}{-} & \multicolumn{1}{c|}{-} & 113M                      \\
\midrule
\multicolumn{1}{l|}{$\text{TRBA}_{PL}$~\cite{baek2021if}}    & \multicolumn{1}{c|}{CVPR2021}            & \multicolumn{1}{c|}{ARD}       & \multicolumn{1}{c|}{Book32~\textit{et al.}}            & 94.8    & 91.3    & 94.0    & 80.6 & 82.7 & 88.1 & -    & -    & -    & -    & \multicolumn{1}{c|}{-} & \multicolumn{1}{c|}{-} & 37M                      \\
\multicolumn{1}{l|}{$\text{ABINet}_{st}$~\cite{abinet}}    & \multicolumn{1}{c|}{CVPR2021}            & \multicolumn{1}{c|}{STD+WiKi}        & \multicolumn{1}{c|}{Uber-Text}               & 96.8    & 94.9    & 97.3    & \underline{87.4} & 90.1 & 93.4 & -    & -    & -    & -    & \multicolumn{1}{c|}{-} & \multicolumn{1}{c|}{-} & 37M                      \\
\multicolumn{1}{l|}{$\text{ABINet}_{est}~\cite{abinet}$}   & \multicolumn{1}{c|}{CVPR2021}             & \multicolumn{1}{c|}{STD+WiKi}       & \multicolumn{1}{c|}{Uber-Text}               & \underline{97.2}    & 95.5    & 97.7    & 86.9 & 89.9 & 94.1 & -    & -    & -    & -    & \multicolumn{1}{c|}{-} & \multicolumn{1}{c|}{-} & 37M                      \\
\multicolumn{1}{l|}{TRBA-cr~\cite{zheng2022pushing}}    & \multicolumn{1}{c|}{CVPR2022}       & \multicolumn{1}{c|}{STD}              & \multicolumn{1}{c|}{ImageNet~\textit{et al.}}          & 97.0    & \underline{96.0}    & \textbf{98.0}    & \textbf{88.8} & 90.9 & \underline{95.1} & \underline{69.9} & \underline{82.3} & \underline{85.3} & 71.9 & \multicolumn{1}{c|}{\underline{85.2}} & \multicolumn{1}{c|}{\underline{80.5}} & 50M                      \\
\midrule
\multicolumn{1}{l|}{Baseline}    & \multicolumn{1}{c|}{-}       & \multicolumn{1}{c|}{STD}                 & \multicolumn{1}{c|}{-}                       & 96.6    & 92.7    & 95.9    & 85.4 & 85.3 & 88.2 & 65.3 & 74.8 & 77.8 & 69.0 & \multicolumn{1}{c|}{79.6} & \multicolumn{1}{c|}{76.2} & 25M \\
\multicolumn{1}{l|}{Ours}        & \multicolumn{1}{c|}{-}   & \multicolumn{1}{c|}{STD}                 & \multicolumn{1}{c|}{URD}                       & \textbf{97.9}    & \textbf{96.4}    & \underline{97.9}    & \textbf{88.8} & \textbf{92.9} & \textbf{95.8} & \textbf{74.2} & \textbf{84.0} & \textbf{85.8} & \textbf{79.4} & \multicolumn{1}{c|}{\textbf{88.2}} & \multicolumn{1}{c|}{\textbf{83.4}} & 25M \\
\bottomrule
\end{tabular}
}
\end{table*}

\subsection{Datasets}
\label{subsec:datasets}

The primary objective of this paper is to exempt the costs associated with human annotation in STR and enhance the practicality of STR methods. This is achieved through the use of labeled synthetic data and unlabeled real data, which eliminates the need for expensive human annotation and improves the performance of STR methods for practical applications. As no manual annotations are required, all experiments are conducted using 100\% of the available data.

\textbf{Synthetic Text Data (STD)} consists of two widely used synthetic datasets, SynthText~\cite{synthtext} and Synth90k~\cite{synth90k}, which respectively contain 8M and 9M images for supervised text recognition.

\textbf{Unlabeled Real Data (URD)} We adopt CC-OCR \cite{DBLP:conf/mm/YangLLWZLTB22}, including 15.77M unlabeled real text images for semi-supervised learning. CC-OCR is cropped with the detection results from Microsoft Azure OCR system on Conceptual Captions Dataset, which covers various real-world scenarios.

\textbf{Scene Text Recognition Benchmarks} Besides the standard benchmarks, we also evaluate our STR models on some challenging STR datasets to demonstrate the effectiveness of our proposed methods in more practical real-world applications. 
These benchmarks cover a wide range of scenarios and vary in fonts, sizes, orientations, and backgrounds. Moreover, some of these benchmarks contain low-quality images with noise, blur, distortion, or occlusion. Therefore, these benchmarks pose great challenges for STR models to recognize diverse and complex text in the wild.
Details are presented in Tab.~\ref{tab:benchmarks}.

\subsection{Implementation Details}
\label{subsec:implementation}
We use ConvNextv2-Nano as the feature extractor for efficiency and performance. To preserve finer features for text recognition, we modify the original strides to $(4,4)$, $(2,1)$, $(2,1)$, $(1,1)$. Therefore, the output feature size is 2 $\times$ 32 for an input image of size 32 $\times$ 128. A two-layer transformer decoder is used for auto-regressive decoding. Each transformer decoder has 8 heads and each head has a dimension of 64.

The hyperparameters for optimization are presented in Tab.~\ref{tab:optim_hyperparams}.

In the Mean-Teacher architecture, two augmented views of an image are taken as input. For the teacher model, a weak augmentation is employed with only color jitters, such as brightness, contrast, and saturation. For the student model, a strong augmentation is applied with both geometry transformations and color jitter, including blur, rotation, crop, perspective distortion, sharpening, and affine transformation.

\subsection{Comparison with State-of-the-Art Methods}
\label{subsec:comp_with_sota}

The performance of the baseline model is significantly improved by our proposed SSL framework, as shown in Tab.~\ref{tab:sota-tr}. Specifically, our entire method achieves an average improvement of 7.2\% over the baseline model. Notably, the improvement is more evident on challenging benchmarks, such as COCO, CTW, TT, HOST, and WOST, where our method surpasses the baseline model by 9.9\%, 9.2\%, 8.0\%, 10.4\%, and 8.6\%, respectively. We believe this is because these challenging datasets have more diverse appearances, which are hard to simulate using synthetic engines.

The ConvnextV2-based baseline model has fewer parameters than some state-of-the-art works, which leads to inferior performance on some benchmarks. However, when combined with our SSL framework, the baseline model can outperform all SOTA methods by a large margin on 11 benchmarks. This demonstrates the robustness of our approach to generalized text recognition and the effectiveness of leveraging unlabeled real images for STR. Moreover, the potential value of unlabeled real images is worth further exploration, as they may contain more useful information for STR than synthetic images.

Our method also outperforms other SSL approaches, including CCR (TRBA-cr) and various self-training methods like $\text{TRBA}_{PL}$, $\text{ABINet}_{st}$, and $\text{ABINet}_{est}$, across nearly all benchmarks. Notably, it achieves an average improvement of nearly \textbf{3.0\%} compared to the leading method, TRBA-cr, highlighting the superiority of our approach over existing semi-supervised STR methods. This improvement is particularly significant in challenging benchmarks such as COCO, CTW, HOST, and WOST, where our method surpasses TRBA-cr by \textbf{4.3\%}, \textbf{1.7\%}, \textbf{7.5\%}, and \textbf{3.0\%}, respectively. We attribute the enhanced performance in complex scenarios like COCO, to its ability to mitigate misalignments in CCR caused by incorrect predictions. By dynamically identifying the most appropriate matching path, our proposed sequential visual consistency effectively reduces these misalignments and enhances overall performance. Furthermore, our sequential semantic consistency strategy demonstrates substantial gains on occluded text datasets such as HOST and WOST. It integrates semantic information into the STR model, significantly improving performance on text instances with incomplete appearances.

\subsection{Ablation on Model Units}
\label{subsec:ablation}
In this section, we analyze the effectiveness of each module. To clearly demonstrate the practicability of the proposed method, besides the average accuracy on all images (termed as Avg\_on\_All), we also report the average results on common (IIIT, SVT, IC13, IC15, SVTP, and CUTE) and more difficult (COCO, CTW, TT, HOST, and WOST) benchmarks, which are termed as Avg\_on\_Com and Avg\_on\_Diff.

We base our method on the state-of-the-art CCR rather than Pseudo Label (PL) and Noisy Student (NS), since \cite{zheng2022pushing} has conducted extensive experiments to exhibit the advantages of CCR.
The results are presented in Tab.~\ref{tab:ablation}. We can observe that each component contributes to the improvement of the baseline model and that our method achieves the best performance with a reasonable combination of these components.
Specifically, the STR model achieves an impressive performance gain of 6\% when only using CCR, confirming the value of unlabeled data. By applying Dynamic Programming, word-level visual consistency regularization further improves the average score by 0.8\%, demonstrating the complementarity of character-level and word-level alignment. Finally, word-level semantic consistency regularization via Reinforcement Learning increases the average result to 83.4\%. Compared to the enhanced CCR (82.2\% \textit{vs.} 80.5\% of original TRBA-cr), our method eventually attains a 1.5\% performance improvement on challenging datasets and an average improvement of 1.2\%. 

To evaluate the consistency of CCR, we employ Cross-Entropy (CE) loss as an alternative to KL-divergence loss. Our experimental results show that CE loss achieves comparable overall recognition performance with KL-divergence loss. Hence, we adopt KL-divergence loss as the default choice in our method.

\begin{figure*}[t]
\centering
\includegraphics[width=1.\linewidth]{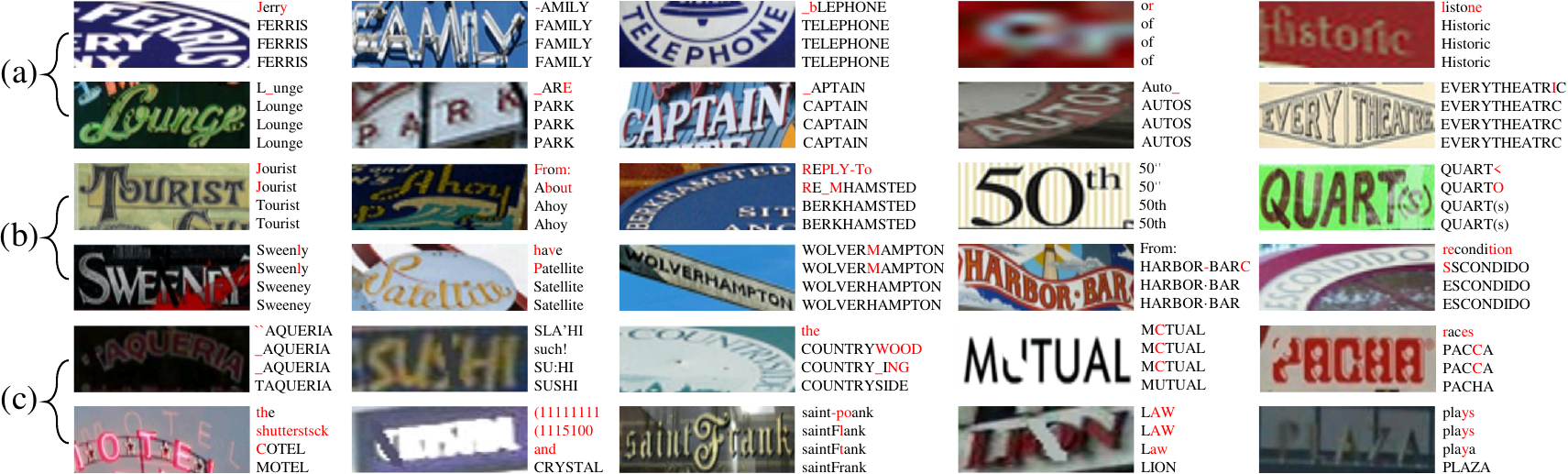}
\caption{Visualization of text recognition results. The four strings near each image represent the prediction of Baseline, Baseline+CCR, Baseline+CCR+WVCR, and Baseline+CCR+WVCR+WSCR respectively. (a), (b), (c) are different image groups to demonstrate the effectiveness of each model unit.}
\label{fig:vis_rec}
\end{figure*}

\subsection{Qualitative Analysis}
\label{subsec:qualitative_analysis}
To demonstrate the effectiveness of our proposed modules, we provide qualitative comparisons in Fig.~\ref{fig:vis_rec}. In group (a), we compare the performance of the STR model trained with and without unlabeled real data. The results show that incorporating unlabeled real data leads to improved performance in various challenging scenarios, such as text with perspective distortion, word art, text disturbed by the background, curved text, and occluded text. This suggests that our proposed module for incorporating unlabeled real data is effective in improving the robustness of the STR model.

In group (b), we compare the performance of the STR model trained with and without Sequential Consistency Regularization (SCR). The results show that SCR enables the STR model to handle even more challenging situations, including more complex backgrounds, smaller characters, heavier perspective distortions, and word art with larger variances. This suggests that SCR is effective in capturing the sequential nature of text and thus improving the consistency of the model's predictions.

As shown in group (c), the STR model without WSCR often misclassifies characters that have similar visual features but different semantic meanings, especially when they are partially occluded. These misclassifications lead to incorrect or meaningless words, which lower the accuracy and readability of the recognition results. By contrast, the STR model with WSCR is more resilient to low-quality images, and can better differentiate these similar characters. Therefore, WSCR can effectively incorporate semantic information into the STR model and thus improves the text recognition performance.

\begin{table}
  \begin{center}
    \caption{Ablation on model units.
      ``CCR'', ``WVCR'' and ``WSCR'' mean character-level CR, word-level visual CR, and word-level semantic CR.}
    \scalebox{0.825}{
      \begin{tabular}{cccccc}
        \toprule
        CCR            & WVCR           & WSCR           & Avg\_on\_Com & Avg\_on\_Diff & Avg\_on\_All  \\
        \midrule
                       &                &                & 92.0 & 70.0 & 76.2 \\
        \cmidrule{4-6}
        \CheckmarkBold &                &                & 94.4 & 77.4 & 82.2 \\
        \cmidrule{4-6}
        \CheckmarkBold & \CheckmarkBold &                & \textbf{94.9} & 78.2 & 83.0 \\
        \cmidrule{4-6}
        \CheckmarkBold & \CheckmarkBold & \CheckmarkBold & \textbf{94.9} & \textbf{78.9} & \textbf{83.4} \\
        \bottomrule
      \end{tabular}
    }
    \label{tab:ablation}
  \end{center}
\end{table}

\section{Conclusion}
\label{sec:conclusion}

    In this paper, we have investigated the use of semi-supervised learning for robust text recognition. Specifically, we propose a novel method that combines character-level consistency regularization with word-level visual and semantic consistency. Our approach takes advantage of multi-granularity and multi-modal alignment, where we align text sequences at different levels of granularity and utilize both visual and semantic information to guide the alignment process.
    We conduct extensive experiments to evaluate the effectiveness of each module in our approach, and our results show that our method outperforms state-of-the-art methods on STR benchmarks by a significant margin. Remarkably, our approach achieves this improvement without relying on human-annotated data, making it more practical in real-world scenarios.
    Our findings demonstrate the potential of semi-supervised learning in enhancing the performance of text recognition models, especially when labeled data is limited or expensive to obtain. In the future, we plan to investigate the application of semi-supervised learning in integrating text recognition with language models, where we can leverage large amounts of unlabeled text data to improve text recognition performance.

\section{Acknowledgments}
Project supported by the Young Scientists Fund of the National Natural Science Foundation of China (Grant No. 62206103).

{\small
\bibliographystyle{ieee}
\bibliography{references}

\begin{thebibliography}{10}\itemsep=-1pt

\bibitem{abuduweili2021adaptive}
A.~Abuduweili, X.~Li, H.~Shi, C.-Z. Xu, and D.~Dou.
\newblock Adaptive consistency regularization for semi-supervised transfer learning.
\newblock In {\em Proc. Conf. Comput. Vision Pattern Recognition}, pages 6923--6932, 2021.

\bibitem{an2023prouda}
P.~An, J.~Liang, T.~Ma, Y.~Chen, L.~Wang, and J.~Ma.
\newblock Prouda: Progressive unsupervised data augmentation for semi-supervised 3d object detection on point cloud.
\newblock {\em Pattern Recognition Letters}, 170:64--69, 2023.

\bibitem{baek2019wrong}
J.~Baek, G.~Kim, J.~Lee, S.~Park, D.~Han, S.~Yun, S.~J. Oh, and H.~Lee.
\newblock What is wrong with scene text recognition model comparisons? dataset and model analysis.
\newblock In {\em Proc. Int. Conf. Comput. Vision}, pages 4715--4723, 2019.

\bibitem{baek2021if}
J.~Baek, Y.~Matsui, and K.~Aizawa.
\newblock What if we only use real datasets for scene text recognition? toward scene text recognition with fewer labels.
\newblock In {\em CVPR}, 2021.

\bibitem{DBLP:journals/access/BaiYLXL18}
X.~Bai, M.~Yang, P.~Lyu, Y.~Xu, and J.~Luo.
\newblock Integrating scene text and visual appearance for fine-grained image classification.
\newblock {\em {IEEE} Access}, 2018.

\bibitem{DBLP:conf/eccv/BautistaA22parseq}
D.~Bautista and R.~Atienza.
\newblock Scene text recognition with permuted autoregressive sequence models.
\newblock In {\em European Conf. Comput. Vision}, 2022.

\bibitem{textistext}
A.~K. Bhunia, A.~Sain, P.~N. Chowdhury, and Y.~Song.
\newblock Text is text, no matter what: Unifying text recognition using knowledge distillation.
\newblock In {\em Proc. Int. Conf. Comput. Vision}, pages 963--972. {IEEE}, 2021.

\bibitem{JVSR}
A.~K. Bhunia, A.~Sain, A.~Kumar, S.~Ghose, P.~N. Chowdhury, and Y.~Song.
\newblock Joint visual semantic reasoning: Multi-stage decoder for text recognition.
\newblock In {\em Proc. Int. Conf. Comput. Vision}, pages 14920--14929, 2021.

\bibitem{bojanowski2017enriching}
P.~Bojanowski, E.~Grave, A.~Joulin, and T.~Mikolov.
\newblock Enriching word vectors with subword information.
\newblock {\em Transactions of the Association for Computational Linguistics}, 5:135--146, 2017.

\bibitem{boschini2022continual}
M.~Boschini, P.~Buzzega, L.~Bonicelli, A.~Porrello, and S.~Calderara.
\newblock Continual semi-supervised learning through contrastive interpolation consistency.
\newblock {\em Pattern Recognition Letters}, 162:9--14, 2022.

\bibitem{CK2017}
C.~K. Chng and C.~S. Chan.
\newblock Total-text: {A} comprehensive dataset for scene text detection and recognition.
\newblock In {\em Proc. ICDAR}, pages 935--942, 2017.

\bibitem{DBLP:conf/eccv/DaWY22levocr}
C.~Da, P.~Wang, and C.~Yao.
\newblock Levenshtein {OCR}.
\newblock In {\em European Conf. Comput. Vision}, volume 13688, pages 322--338, 2022.

\bibitem{DBLP:conf/ijcai/DuCJYZLDJ22svtr}
Y.~Du, Z.~Chen, C.~Jia, X.~Yin, T.~Zheng, C.~Li, Y.~Du, and Y.~Jiang.
\newblock {SVTR:} scene text recognition with a single visual model.
\newblock In L.~D. Raedt, editor, {\em Int. Joint Conf. on Artificial Intelligence}, pages 884--890, 2022.

\bibitem{abinet}
S.~Fang, H.~Xie, Y.~Wang, Z.~Mao, and Y.~Zhang.
\newblock Read like humans: Autonomous, bidirectional and iterative language modeling for scene text recognition.
\newblock In {\em CVPR}, 2021.

\bibitem{feng2023weakly}
J.~Feng, X.~Wang, T.~Li, S.~Ji, and W.~Liu.
\newblock Weakly-supervised semantic segmentation via online pseudo-mask correcting.
\newblock {\em Pattern Recognition Letters}, 165:33--38, 2023.

\bibitem{gu2018neural}
J.~Gu, D.~J. Im, and V.~O. Li.
\newblock Neural machine translation with gumbel-greedy decoding.
\newblock In {\em AAAI Conf. on Artificial Intelligence}, volume~32, 2018.

\bibitem{guan2023self}
T.~Guan, C.~Gu, J.~Tu, X.~Yang, Q.~Feng, Y.~Zhao, and W.~Shen.
\newblock Self-supervised implicit glyph attention for text recognition.
\newblock In {\em CVPR}, 2023.

\bibitem{DBLP:conf/icml/GuoPSW17calib}
C.~Guo, G.~Pleiss, Y.~Sun, and K.~Q. Weinberger.
\newblock On calibration of modern neural networks.
\newblock In {\em Int. Conf. Mach. Learning}, pages 1321--1330, 2017.

\bibitem{synthtext}
A.~Gupta, A.~Vedaldi, and A.~Zisserman.
\newblock Synthetic data for text localisation in natural images.
\newblock In {\em CVPR}, 2016.

\bibitem{DBLP:conf/aaai/HeC0LHWD22sgtr}
Y.~He, C.~Chen, J.~Zhang, J.~Liu, F.~He, C.~Wang, and B.~Du.
\newblock Visual semantics allow for textual reasoning better in scene text recognition.
\newblock In {\em AAAI Conf. on Artificial Intelligence}, pages 888--896, 2022.

\bibitem{synth90k}
M.~Jaderberg, K.~Simonyan, A.~Vedaldi, and A.~Zisserman.
\newblock Synthetic data and artificial neural networks for natural scene text recognition.
\newblock {\em CoRR}, abs/1406.2227, 2014.

\bibitem{DBLP:conf/iclr/JangGP17}
E.~Jang, S.~Gu, and B.~Poole.
\newblock Categorical reparameterization with gumbel-softmax.
\newblock In {\em Int. Conf. on Learning Representations}, 2017.

\bibitem{karatzas2015icdar}
D.~Karatzas, L.~Gomez{-}Bigorda, A.~Nicolaou, S.~K. Ghosh, A.~D. Bagdanov, M.~Iwamura, J.~Matas, L.~Neumann, V.~R. Chandrasekhar, S.~Lu, F.~Shafait, S.~Uchida, and E.~Valveny.
\newblock {ICDAR} 2015 competition on robust reading.
\newblock In {\em Proc. ICDAR}, pages 1156--1160, 2015.

\bibitem{karatzas2013icdar}
D.~Karatzas, F.~Shafait, S.~Uchida, M.~Iwamura, L.~G. i~Bigorda, S.~R. Mestre, J.~Mas, D.~F. Mota, J.~A. Almazan, and L.~P. de~las Heras.
\newblock Icdar 2013 robust reading competition.
\newblock In {\em Proc. ICDAR}, pages 1484--1493, 2013.

\bibitem{satrn}
J.~Lee, S.~Park, J.~Baek, S.~J. Oh, S.~Kim, and H.~Lee.
\newblock On recognizing texts of arbitrary shapes with 2d self-attention.
\newblock In {\em Proc. Conf. Comput. Vision Pattern Recognition Workshops}, pages 2326--2335, 2020.

\bibitem{LiaoLHYWB21}
M.~Liao, P.~Lyu, M.~He, C.~Yao, W.~Wu, and X.~Bai.
\newblock Mask textspotter: An end-to-end trainable neural network for spotting text with arbitrary shapes.
\newblock {\em IEEE Trans. Pattern Anal. Mach. Intell.}, 43(2):532--548, 2021.

\bibitem{ctw1500}
Y.~Liu, L.~Jin, S.~Zhang, C.~Luo, and S.~Zhang.
\newblock Curved scene text detection via transverse and longitudinal sequence connection.
\newblock {\em Pattern Recognition}, 90:337--345, 2019.

\bibitem{cluo2019moran}
C.~Luo, L.~Jin, and Z.~Sun.
\newblock Moran: A multi-object rectified attention network for scene text recognition.
\newblock {\em Pattern Recognition}, 90:109--118, 2019.

\bibitem{wiki}
S.~Merity, C.~Xiong, J.~Bradbury, and R.~Socher.
\newblock Pointer sentinel mixture models.
\newblock In {\em Int. Conf. on Learning Representations}, 2017.

\bibitem{DBLP:conf/cvpr/MishraAJ12}
A.~Mishra, K.~Alahari, and C.~V. Jawahar.
\newblock Top-down and bottom-up cues for scene text recognition.
\newblock In {\em Proc. Conf. Comput. Vision Pattern Recognition}, 2012.

\bibitem{patel2023seq}
G.~Patel, J.~P. Allebach, and Q.~Qiu.
\newblock Seq-ups: Sequential uncertainty-aware pseudo-label selection for semi-supervised text recognition.
\newblock In {\em Winter Conf. on App. of Comput. Vision}, 2023.

\bibitem{quy2013recognizing}
T.~Quy~Phan, P.~Shivakumara, S.~Tian, and C.~Lim~Tan.
\newblock Recognizing text with perspective distortion in natural scenes.
\newblock In {\em ICCV}, pages 569--576, 2013.

\bibitem{conf/cvpr/RennieMMRG17}
S.~J. Rennie, E.~Marcheret, Y.~Mroueh, J.~Ross, and V.~Goel.
\newblock Self-critical sequence training for image captioning.
\newblock In {\em CVPR}, 2017.

\bibitem{risnumawan2014robust}
A.~Risnumawan, P.~Shivakumara, C.~S. Chan, and C.~L. Tan.
\newblock A robust arbitrary text detection system for natural scene images.
\newblock {\em Expert Syst. Appl.}, 41(18):8027--8048, 2014.

\bibitem{RizveDRS21semi}
M.~N. Rizve, K.~Duarte, Y.~S. Rawat, and M.~Shah.
\newblock In defense of pseudo-labeling: An uncertainty-aware pseudo-label selection framework for semi-supervised learning.
\newblock In {\em ICLR}, 2021.

\bibitem{crnn}
B.~Shi, X.~Bai, and C.~Yao.
\newblock An end-to-end trainable neural network for image-based sequence recognition and its application to scene text recognition.
\newblock {\em IEEE Trans. Pattern Anal. Mach. Intell.}, 39(11):2298--2304, 2017.

\bibitem{aster}
B.~Shi, M.~Yang, X.~Wang, P.~Lyu, C.~Yao, and X.~Bai.
\newblock {ASTER:} an attentional scene text recognizer with flexible rectification.
\newblock {\em IEEE Trans. Pattern Anal. Mach. Intell.}, 41(9):2035--2048, 2019.

\bibitem{sohn2020fixmatch}
K.~Sohn, D.~Berthelot, N.~Carlini, Z.~Zhang, H.~Zhang, C.~A. Raffel, E.~D. Cubuk, A.~Kurakin, and C.-L. Li.
\newblock Fixmatch: Simplifying semi-supervised learning with consistency and confidence.
\newblock In {\em NIPS}, 2020.

\bibitem{DBLP:conf/nips/TarvainenV17meanteacher}
A.~Tarvainen and H.~Valpola.
\newblock Mean teachers are better role models: Weight-averaged consistency targets improve semi-supervised deep learning results.
\newblock In {\em Neural Inform. Process. Syst.}, pages 1195--1204, 2017.

\bibitem{DBLP:conf/nips/VaswaniSPUJGKP17attn}
A.~Vaswani, N.~Shazeer, N.~Parmar, J.~Uszkoreit, L.~Jones, A.~N. Gomez, L.~Kaiser, and I.~Polosukhin.
\newblock Attention is all you need.
\newblock In {\em Neural Inform. Process. Syst.}, pages 5998--6008, 2017.

\bibitem{cocotext}
A.~Veit, T.~Matera, L.~Neumann, J.~Matas, and S.~J. Belongie.
\newblock Coco-text: Dataset and benchmark for text detection and recognition in natural images.
\newblock {\em CoRR}, abs/1601.07140, 2016.

\bibitem{wang2011end}
K.~Wang, B.~Babenko, and S.~Belongie.
\newblock End-to-end scene text recognition.
\newblock In {\em Proc. Int. Conf. Comput. Vision}, pages 1457--1464, 2011.

\bibitem{DBLP:conf/eccv/WangDY22multigra}
P.~Wang, C.~Da, and C.~Yao.
\newblock Multi-granularity prediction for scene text recognition.
\newblock In {\em European Conf. Comput. Vision}, 2022.

\bibitem{ost}
Y.~Wang, H.~Xie, S.~Fang, J.~Wang, S.~Zhu, and Y.~Zhang.
\newblock From two to one: {A} new scene text recognizer with visual language modeling network.
\newblock In {\em Proc. Int. Conf. Comput. Vision}, pages 14174--14183, 2021.

\bibitem{woo2023convnext}
S.~Woo, S.~Debnath, R.~Hu, X.~Chen, Z.~Liu, I.~S. Kweon, and S.~Xie.
\newblock Convnext v2: Co-designing and scaling convnets with masked autoencoders.
\newblock In {\em Proc. Conf. Comput. Vision Pattern Recognition}, 2023.

\bibitem{conf/nips/XieDHL020}
Q.~Xie, Z.~Dai, E.~H. Hovy, T.~Luong, and Q.~Le.
\newblock Unsupervised data augmentation for consistency training.
\newblock In {\em Neural Inform. Process. Syst.}, 2020.

\bibitem{DBLP:conf/eccv/XieFZWB22wordart}
X.~Xie, L.~Fu, Z.~Zhang, Z.~Wang, and X.~Bai.
\newblock Toward understanding wordart: Corner-guided transformer for scene text recognition.
\newblock In {\em European Conf. Comput. Vision}, volume 13688, pages 303--321, 2022.

\bibitem{Pren2D}
R.~Yan, L.~Peng, S.~Xiao, and G.~Yao.
\newblock Primitive representation learning for scene text recognition.
\newblock In {\em CVPR}, 2021.

\bibitem{YangGLHBBYB19}
M.~Yang, Y.~Guan, M.~Liao, X.~He, K.~Bian, S.~Bai, C.~Yao, and X.~Bai.
\newblock Symmetry-constrained rectification network for scene text recognition.
\newblock In {\em Proc. Int. Conf. Comput. Vision}, pages 9146--9155, 2019.

\bibitem{DBLP:conf/mm/YangLLWZLTB22}
M.~Yang, M.~Liao, P.~Lu, J.~Wang, S.~Zhu, H.~Luo, Q.~Tian, and X.~Bai.
\newblock Reading and writing: Discriminative and generative modeling for self-supervised text recognition.
\newblock In {\em Int. Conf. Multimedia}, 2022.

\bibitem{zhang2020sahan}
J.~Zhang, C.~Luo, L.~Jin, T.~Wang, Z.~Li, and W.~Zhou.
\newblock Sahan: Scale-aware hierarchical attention network for scene text recognition.
\newblock {\em Pattern Recognition Letters}, 2020.

\bibitem{DBLP:conf/aaai/ZhangZYSL022conclr}
X.~Zhang, B.~Zhu, X.~Yao, Q.~Sun, R.~Li, and B.~Yu.
\newblock Context-based contrastive learning for scene text recognition.
\newblock In {\em AAAI}, 2022.

\bibitem{zhang2021pmmn}
Y.~Zhang, Z.~Fu, F.~Huang, and Y.~Liu.
\newblock Pmmn: pre-trained multi-modal network for scene text recognition.
\newblock {\em Pattern Recognition Letters}, 2021.

\bibitem{zheng2022pushing}
C.~Zheng, H.~Li, S.-M. Rhee, S.~Han, J.-J. Han, and P.~Wang.
\newblock Pushing the performance limit of scene text recognizer without human annotation.
\newblock In {\em Proc. Conf. Comput. Vision Pattern Recognition}, pages 14116--14125, 2022.

\bibitem{zhu2017cascaded}
Y.~Zhu, M.~Liao, M.~Yang, and W.~Liu.
\newblock Cascaded segmentation-detection networks for text-based traffic sign detection.
\newblock {\em IEEE transactions on intelligent transportation systems}, 19(1):209--219, 2017.

\end{thebibliography}
}

\end{document}